\documentclass{article}

\PassOptionsToPackage{numbers, compress}{natbib}

 \usepackage[preprint]{nips_2018}

\usepackage[utf8]{inputenc} 
\usepackage[T1]{fontenc}    
\usepackage{hyperref}       
\usepackage{url}            
\usepackage{booktabs}       
\usepackage{amsfonts}       
\usepackage{nicefrac}       
\usepackage{microtype}      
\usepackage[cmex10]{amsmath}
\usepackage{amssymb}
\usepackage{graphicx}
\usepackage{subfig}

\title{How Much Can We Really Trust You? Towards Simple, Interpretable Trust Quantification Metrics for Deep Neural Networks}

\author{
  Alexander Wong$^{1,2,3,*}$, Xiao Yu Wang$^{1,3}$, and Andrew Hryniowski$^{1,2,3}$\\
  $^{1}$ Vision and Image Processing Research Group, University of Waterloo, Waterloo, ON, Canada\\
  $^{2}$ Waterloo Artificial Intelligence Institute, University of Waterloo, Waterloo, ON, Canada\\
  $^{3}$ DarwinAI Corp., Waterloo, ON, Canada \\
  \texttt{$^{*}$ a28wong@uwaterloo.ca} \\
}

\begin{document}

\maketitle

\begin{abstract}
A critical step to building trustworthy deep neural networks is trust quantification, where we ask the question: \textbf{How much can we trust a deep neural network?}  In this study, we take a step towards simple, interpretable metrics for trust quantification by introducing a suite of metrics for assessing the overall trustworthiness of deep neural networks based on their behaviour when answering a set of questions.  We conduct a thought experiment and explore two key questions about trust in relation to confidence: 1) How much trust do we have in actors who give wrong answers with great confidence?, and 2) How much trust do we have in actors who give right answers hesitantly?  Based on insights gained, we introduce the concept of \textbf{question-answer trust} to quantify trustworthiness of an individual answer based on confident behaviour under correct and incorrect answer scenarios, and the concept of \textbf{trust density} to characterize the distribution of overall trust for an individual answer scenario. We further introduce the concept of \textbf{trust spectrum} for representing overall trust with respect to the spectrum of possible answer scenarios across correctly and incorrectly answered questions.  Finally, we introduce \textbf{NetTrustScore}, a scalar metric summarizing overall trustworthiness.  The suite of metrics aligns with past social psychology studies that study the relationship between trust and confidence.  Leveraging these metrics, we quantify the trustworthiness of several well-known deep neural network architectures for image recognition to get a deeper understanding of where trust breaks down.  The proposed metrics are by no means perfect, but the hope is to push the conversation towards better metrics to help guide practitioners and regulators in producing, deploying, and certifying deep learning solutions that can be trusted to operate in real-world, mission-critical scenarios.
\end{abstract}

\section{Introduction}
\label{Introduction}

Deep learning~\cite{lecun2015deep} has led to the recent revolution in artificial intelligence across numerous fields, ranging from medicine~\cite{wallach2015atomnet,alperstein2019smiles,wang2020covidnet,7158331} and physical sciences~\cite{Torlai,Guest_2018}, to language~\cite{vaswani2017attention,devlin2018bert,brown2020language} and music~\cite{briot2017deep}, to speech~\cite{hannun2014deep,xiong2016achieving} and vision~\cite{krizhevsky2012imagenet,ResNet50,inceptionv3,chen2017rethinking}.  As we move towards the widespread deployment of deep learning as a ubiquitous technology across industries and society, one question keeps coming up in debates and conversations: \textbf{Can we trust deep neural networks?}  This question has become a challenge in real-world deployment of deep neural networks, particularly for mission-critical and regulated spaces such as automotive, finance, and healthcare.  As such, it is important to tackle the challenge of trustworthiness in the context of deep learning given its importance for facilitating widespread, real-world adoption of deep learning as the solution for a wide range of complex problems across industries and societies that can benefit the global population.  A critical step to building trustworthy deep neural networks is trust quantification, where the goal is to quantitatively assess the trustworthiness of deep neural networks in the decisions that they make.  Therefore, here we are instead asking the following question: \textbf{How much can we trust a deep neural network?}

While there exists a wealth of literature in the quantification of other aspects of deep learning such as explainability~\cite{lin2019explanations}, efficiency~\cite{canziani2016analysis,Wong2018_Netscore}, and robustness~\cite{huang2016safety,8418593,aboutalebi2020vulnerability}, the quantification of trustworthiness of deep neural networks is a very new area of interest and not well explored.  Much of the literature related to trust quantification have focused on quantifying the trustworthiness of individual predictions made by deep neural networks for a single data sample.  For example, one direction that has been explored relates to uncertainty estimation of deep learning networks~\cite{titensky2018uncertainty,geifman2018biasreduced,NIPS2017_7141,gal2015dropout}, where the trustworthiness of individual predictions made by deep neural networks may be quantified by treating high uncertainty as low prediction trustworthiness.  A limitation with a number of previously proposed uncertainty estimation approaches lies in high complexity and limited interpretability, with many returning distributions over the space of possible predictions and limited to only Bayesian neural networks~\cite{NIPS2017_7141,gal2015dropout}.  Furthermore, such approaches do not take into account the correctness of the predictions made by the deep neural network, thus making it difficult to assess trustworthiness using such approaches in a way that is backed by data-driven evidence.  Finally, to the best of the authors' knowledge, the quantification of overall trustworthiness of a deep neural network as a scalar metric via uncertainty estimation has not been previously explored.

Another direction is an agreement-based approach proposed by Jiang et al.~\cite{jiang2018trust}, where the trustworthiness of an individual prediction for a single data sample is quantified based on the agreement between the classifier and a modified nearest-neighbor classifier that data sample.  While more broadly applicable beyond deep neural networks, such an approach is highly complex with high computational complexity given the considerable number of distance measurements needed, and also does not provide an assessment of the overall trustworthiness of the classifier.  More recently, a subjective logic inspired approach was proposed~\cite{deeptrust} for constructing a probabilistic logic description of deep neural networks and producing trust probabilities around a neural network's prediction.  A key limitation of this approach to trust quantification, as indicated by the authors, is the dependency on the opinions of data and as such requires trust information about data.  Furthermore, the exploration of this approach was limited to smaller multi-layer perceptron networks.

In this study, we take a step towards simple, interpretable metrics for trust quantification by introducing a suite of metrics for assessing the overall trustworthiness of deep neural networks based on their behaviour when answering a set of questions, namely:
\begin{itemize}
    \item \textbf{Question-answer trust}: a quantification of trustworthiness of an individual answer given by a deep neural network based on its behaviour under correct and incorrect answer scenarios,
    \item \textbf{Trust density}: a characterization of the distribution of overall trust of a deep neural network for an individual answer scenario,
    \item \textbf{Trust spectrum}: a representation of overall trust with respect to the spectrum of possible answer scenarios  across both correctly and incorrectly answered questions, and
    \item \textbf{NetTrustScore}: a scalar metric summarizing the overall trustworthiness of a deep neural network based on the trust spectrum.
\end{itemize}

The paper is organized as follows. In Section~\ref{Methods}, a detailed description of the underlying theory behind question-answer trust, trust density, trust spectrum, NetTrustScore, and related expected confidence metrics is presented.  In Section~\ref{Results}, experimental results are presented where we leveraged the proposed suite of metrics experimentally to quantify the overall trustworthiness of several well-known, complex deep neural network architectures for the task of image recognition at different levels of granularity to get a deeper understanding of where trust breaks down.  Conclusions are drawn and future work discussed in Section~\ref{Conclusions}.  The broader impact of this work is discussed in Section~\ref{impact}.

\section{Methods}
\label{Methods}
Here, we will describe a suite of metrics for assessing the overall trustworthiness of deep neural networks based on their behaviour when answering a set of questions.  Our main goal is to devise metrics with two main properties in mind:
\begin{itemize}
\item \textbf{Simple to compute}: The metric should be simple to compute to enable not only one-off, end-of-cycle trust quantification, but also regular trust quantification cycles to facilitate continuous monitoring of trust behaviour to identify when a deep neural network is not longer trustworthy due to changing data dynamics.
\item \textbf{Intepretable}: The metric should be interpretable to a human end-user to allow for greater transparency to how it assesses the degree of trustworthiness of a deep neural network.
\end{itemize}

In an attempt to devise metrics that satisfy these two properties, let us first conduct a thought experiment and explore two key philosophical questions about trust in relation to confidence to help guide our definition of trustworthiness:
\begin{enumerate}
\item How much trust do we have in actors who give wrong answers with great confidence?
\item How much trust do we have in actors who give right answers hesitantly?
\end{enumerate}

\begin{figure}
\centering
	\includegraphics[width = 1\linewidth]{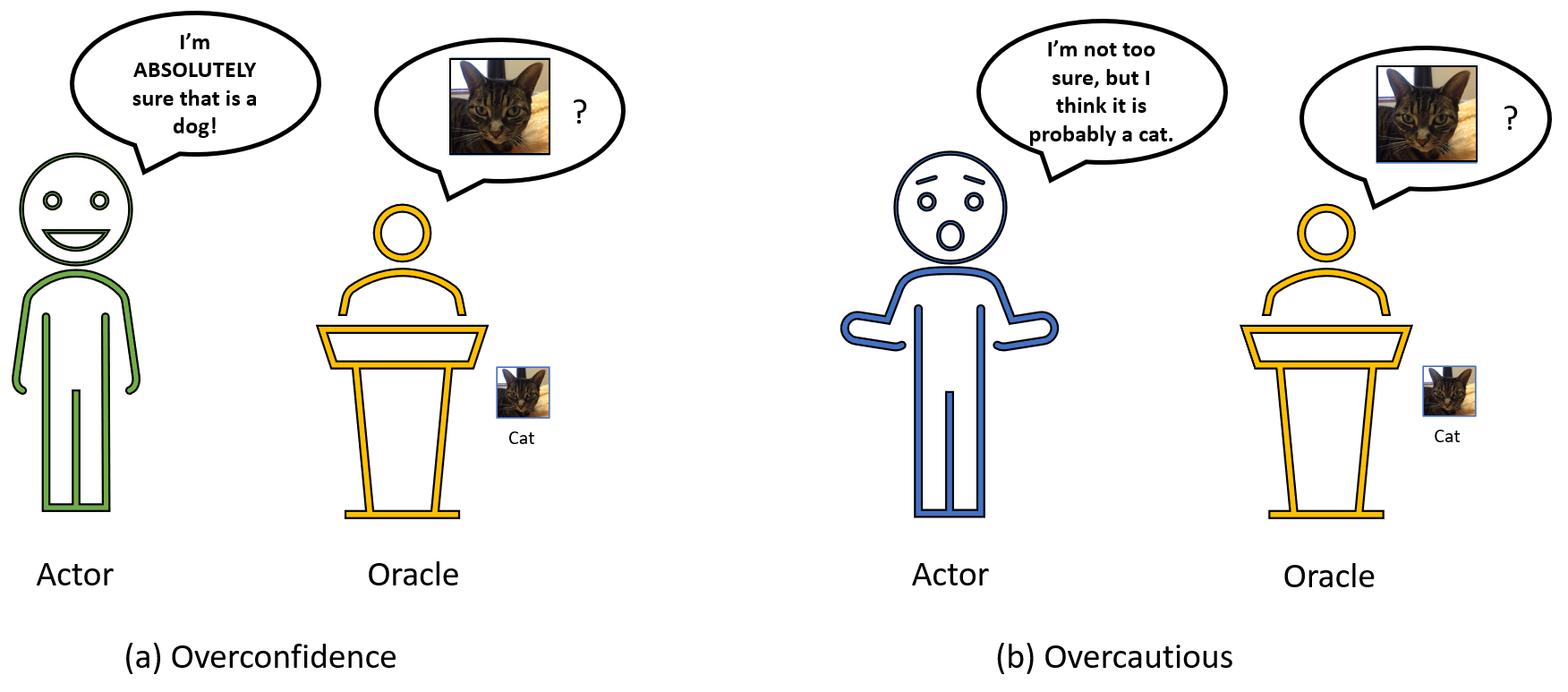}
	\caption{Thought experiment where we explore two philosophical questions in relation to confidence to guide our definition of trustworthiness: (a) \textbf{Overconfidence}: How much trust do we have in actors who give wrong answers with great confidence?, and (b) \textbf{Overcautious}: How much trust do we have in actors who give right answers hesitantly? We explore this from the perspective of an oracle, who knows all the answers and can judge the correctness of an actor's answers in an unbiased manner.}	
	\label{fig:confidence}
\vspace{-0.21 cm}
\end{figure}

We explore these two questions from the perspective of an oracle, who knows all the answers ahead of time and can judge the correctness of an actor's answers in an unbiased manner.  Under this lens, we make the following logical assumptions of trustworthiness:
\begin{itemize}
\item \textbf{Overconfidence}: The more confident an actor is about their wrong answer, the less trust one has in the actor.
\item \textbf{Overcautious}: The less confident an actor is about their right answer, the less trust one has in the actor.
\end{itemize}
These logical assumptions align with a recent social psychology study conducted by Tenney et al.~\cite{Tenney} where participants assessed targets who were either overconfident or cautious, and subsequently learned of the target's actual performance.  It was found in that study that targets who were overconfident (high confidence coupled with poor performance) were evaluated less positively by the participants and avoided as collaborators (which can be interpreted as less trustworthy), particularly for falsified claims.  It was also found that targets who were cautious were evaluated less positively than targets who were more confident.  Furthermore, other studies~\cite{Tenney07,Tenney08} have found that individuals who were more confident and later revealed to be wrong (thus shown to be overconfident based on knowledge of poor performance) were less believable in the eyes of others.  As such, based on these logical assumptions about trustworthiness in relation to confidence, one can quantify the overall trustworthiness of an actor by giving the actor a set of test questions and observing the actor's confidence when answering questions correctly and incorrectly.

\subsection{Question-Answer Trust}
Based on the insights from the thought experiment as well as the desired properties we wish the metrics to possess, we can now attempt to formulate metrics for quantifying the overall trustworthiness of a deep neural network based on its behaviour when answering questions correctly and incorrectly. Let us define the relationship for an question-answer pair $(x,y)$ with respect to model $M$ as
\begin{equation}
y = M(x)
\end{equation}
\noindent where $x \in X$ denotes the question, $X$ is the space of all possible questions, $M$ denotes the model, $y \in Z$ denotes the answer, and $Z$ is the space of all possible answers.  Given an oracle $O$, let $z \in Z$ denote the oracle answer corresponding to question $x$,
\begin{equation}
z = O(x)
\end{equation}

We can now define $R_{y \neq z|M}$ denote the space of all questions $x$ where the answer $y$ by model $M$ does not match the oracle answer $z$ (i.e., incorrect answers), and $R_{y = z|M}$ denote the space of all questions $x$ where the answer $y$ by model $M$ matches the oracle answer $z$ (i.e., correct answers).  Finally, let us define the confidence of $M$ in an answer $y$ to question $x$ as $C(y|x)$.

Given the assumptions that the trustworthiness of an actor increases as confidence increases when giving correct answers, and decreases as actor confidence increases when giving incorrect answers, we first define the concept of \textbf{question-answer trust}, a quantification of trustworthiness of an answer $y$ given by model $M$ for a given question $x$, with full knowledge of the oracle answer $z$.  More specifically, we define the question-answer trust $Q_{z}(x,y)$ for a given question-answer pair $(x,y)$ as
\begin{equation}
    Q_{z}(x,y) =
    \begin{cases}
      C(y|x)^\alpha & \text{if $x \in R_{y = z|M}$ } \\
      (1 - C(y|x))^\beta & \text{if $x \in R_{y \neq z|M}$ } \\
    \end{cases}
    \label{questiontrust}
\end{equation}
\noindent where $\alpha$ and $\beta$ denote reward and penalty relaxation coefficients.

From an interpretation perspective, the proposed question-answer trust expressed in Equation~\ref{questiontrust} quantifies how well placed the deep neural network's confidence is on its answer.  The more well placed its confidence is in both correct and incorrect answer scenarios, the higher question-answer trust is and the more trustworthiness the deep neural network is for that particular answer.  Based on how well-placed its confidence is expected to be under both correct and incorrect scenarios, the proposed question-answer trust fundamentally accomplishes the following:
\begin{itemize}
\item \textbf{Rewards Well-Placed Confidence}: The first case in Equation~\ref{questiontrust} rewards the level of confidence of model $M$ in a scenario where the question is correctly answered, with the amount of reward controlled by $\alpha$.  As such, a deep neural network that has higher confidence when answering correctly will receive a higher question-answer trust for that question.
\item \textbf{Penalizes Undeserved Overconfidence}:  The second case in Equation~\ref{questiontrust} penalizes how much the model $M$ is  overconfident in a scenario where the question is incorrectly answered, with the amount of penalty controlled by $\beta$.  A deep neural network that has higher confidence when answering incorrectly will receive a lower question-answer trust for that question.
\end{itemize}

\subsection{Trust Density}
\begin{figure}
\centering
	\includegraphics[width = 0.6\linewidth]{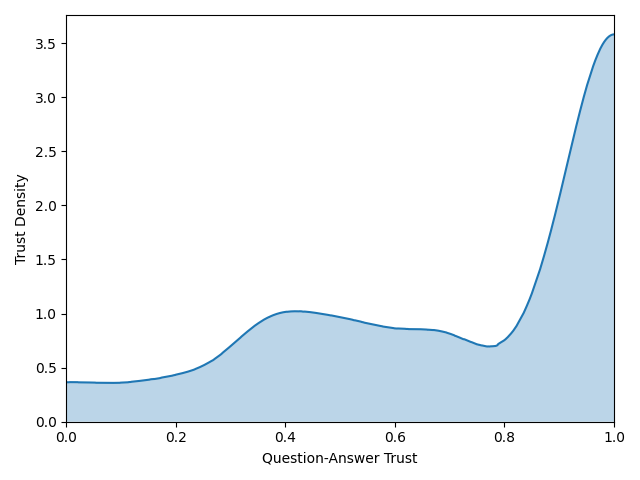}
	\caption{The trust density of a ResNet-50 deep neural network on a subset of the ImageNet~\cite{ImageNet} dataset for a specific answer scenario (in this case, ``water bottle''). A trust density characterizes the distribution of overall trust of a deep neural network for an individual answer scenario.}	
	\label{fig:trustdensity_resnet50}
\vspace{-0.21 cm}
\end{figure}

Based on the notion of question-answer trust $Q_z(x,y)$, we now introduce the concept of \textbf{trust density} as a quantitative means to study the overall trust behaviour of a deep neural network for an individual answer scenario.  More specifically, a trust density $F(Q_z)$ represents the distribution of question-answer trust $Q_z(x,y)$ for all questions $x$ that are answered as a given answer $z$. As an example, the trust density of a ResNet-50 deep neural network for the task of image recognition on a subset of the ImageNet~\cite{ImageNet} dataset for a specific answer scenario (in this case, ``water bottle'') is shown in Figure~\ref{fig:trustdensity_resnet50}.  Note that the trust densities shown in this study, given the finite number of questions, are approximations of true underlying trust density. These approximations are computed via non-parametric density estimation with a Gaussian kernel of bandwidth $\frac{\gamma}{\sqrt{N}}$ where $\gamma_j$ is a kernel constant. In this study, $\gamma$ is set to $0.5$. A reflection condition is used for boundary events.

\subsection{Trust Spectrum}
\begin{figure}
\centering
	\includegraphics[width = \linewidth]{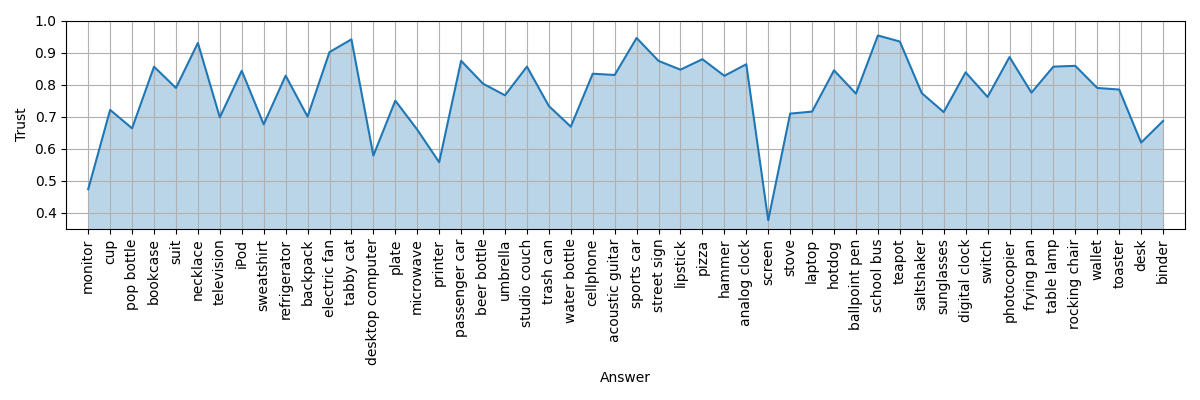}
	\caption{The trust spectrum of ResNet-50 on a subset of the ImageNet~\cite{ImageNet} dataset.  A trust spectrum represents overall trust with respect to the spectrum of possible answer scenarios based on confident behaviour across correctly and incorrectly answered questions.}	
	\label{fig:trustspectrum_resnet50}
\vspace{-0.21 cm}
\end{figure}

To analyze where the trust of a deep neural network may break down at a network level, we now introduce the concept of \textbf{trust spectrum} to study the overall trust behaviour across all possible answer scenarios.  More specifically, a trust spectrum is a function representing overall trust with respect to the spectrum of possible answer scenarios based on question-answer trust across correctly and incorrectly answered questions. Let the trust spectrum be defined as $\{T_M(z)\}_{z \in Z}$, where $T_M(z)$ is the trust spectrum coefficient for an answer scenario $z$. We define $T_M(z)$ as the question-answer trust integral across all question-answer pairs $(x,y)$,
\begin{equation}
T_M\left(z\right) = \int \int P(x,y)Q_z(x,y) dydx
\label{spectrum}
\end{equation}
\noindent where $P(x,y)$ is the probability of the occurrence of question-answer pair $(x,y)$.  In this study, we assume a uniform and independent distribution for the probability of occurrence $P(x,y)$. For a deterministic and differentiable model $M$ with uniform  distribution, Equation~\ref{spectrum} reduces to
\begin{equation}
T_M\left(z\right) = \frac{1}{N} \int Q_z(x) dx
\label{spectrum_reduced}
\end{equation}
where $N$ is the number of questions. Analyzing the trust spectrum $\{T_M(z)\}_{z \in Z}$ can provide valuable, in-depth insights into where trust can break down.  These insights can then be leveraged to either improve the deep neural network or to avoid its usage for the most untrustworthy types of scenarios.  As an example, the trust spectrum of a ResNet-50 deep neural network for the task of image recognition on a subset of the ImageNet~\cite{ImageNet} dataset is shown in Figure~\ref{fig:trustspectrum_resnet50}.

\subsection{NetTrustScore}
To summarize the overall trustworthiness of a deep neural network based on the information within the trust spectrum $\{T_M(z)\}_{z \in Z}$ into a single scalar metric, we define \textbf{NetTrustScore} $T_M$ as the expectation of $T_M(z)$ across all possible answer scenarios $z$,

\begin{equation}
T_M = \int P(z)T_M(z) dz
\label{trustscore}
\end{equation}
\noindent where $P(z)$ is the probability of occurrence of answer scenario $z$.

From an interpretation perspective, the proposed NetTrustScore is fundamentally a quantitative score that indicates how well placed the deep neural network's confidence is expected to be under all possible answer scenarios that can occur.  This makes NetTrustScore particularly useful for conducting a high-level quantitative comparative analysis across different deep neural networks to quickly identify a set of candidates with the highest overall trustworthiness for potential deployment and further analysis at finer granularity using trust spectrum and trust density.

\subsection{Implementation Considerations}
In this study, we set $\alpha=1$ and $\beta=1$ to penalize undeserved overconfidence and reward well-placed confidence equally.  However, depending on the particular application in question for employing the proposed suite of metrics as means for trust quantification, one may adjust the reward and penalty relaxation coefficients based on the desired trade-off and goal at hand.  For example, in the case of clinical decision support where incorrect diagnostic predictions or inappropriate treatment recommendations that can negatively impact patient health, one may increase $\beta$ to impose greater penalties on overconfident behaviour of a deep neural network.  As a final point to consider, it can be seen from the formulation of the aforementioned question-answer trust, trust density, trust spectrum, and NetTrustScore that the proposed suite of metrics are simple to calculate in practice across a test dataset in a discrete form.

\subsection{Expected Confidence}

Finally, as complementary scalar metrics to NetTrustScore, we further define the expected confidence for correct answers ${\bar C}_{y = z|x}$ and the expected confidence for incorrect answers ${\bar C}_{y \neq z|x,M}$ to gain a better understanding behind the behaviour of deep neural networks when giving correct answers and incorrect answers,

\begin{equation}
{\bar C}_{y = z} = \int_{y \in R_{y = z|M}}P(x,y)C(y|x)
\label{confidence_correct}
\end{equation}

\begin{equation}
{\bar C}_{y \neq z} = \int_{y \in R_{y \neq z|M}}P(x,y)C(y|x)
\label{confidence_correct}
\end{equation}
\noindent where $P(x)$ is the probability of the occurrence of question-answer pair $(x,y)$.  In this study, we assume a uniform distribution for the probability of occurrence.

\section{Results and Discussion}
\label{Results}
\begin{table}[h]
	\centering
	\caption{NetTrustScore ($T_M$), expected confidence for correct answers (${\bar C}_{y = z}$), expected confidence for incorrect answers (${\bar C}_{y \neq z}$),  the percentage of correct answers ($n_{y=z}$), percentage of incorrect answers ($n_{y \neq z}$), and number of parameters for four deep image recognition networks (MobileNet-V1~\cite{MobileNetv1}, MobileNet-V2~\cite{MobileNetv2}, ShuffleNet-V2~\cite{ShuffleNetv2}, and ResNet-50~\cite{ResNet50}.   }
	\begin{tabular}{p{3cm}|ccccc|c}
		\hline
		Model ($M$)  & NetTrustScore ($T_M$) & $n_{y=z}$&	${\bar C}_{y = z}$ & $n_{y\neq z}$ & ${\bar C}_{y \neq z}$ & params (M)\\
		\hline 
		MobileNet-V1~\cite{MobileNetv1}	&	0.713	&$64.5\%$	& 0.839 &$35.5\%$& 0.515 & 3.26\\
		ShuffleNet-V2~\cite{ShuffleNetv2} 	&	0.723&$66.1\%$	&	0.839 &$33.9\%$& 0.502 & 1.32\\
		MobileNet-V2~\cite{MobileNetv2}		&	0.739&$68.7\%$	&	0.845 &$31.3\%$& 0.493 & 2.29\\		
		ResNet-50~\cite{ResNet50} 	&	0.776 &$75.6\%$	&	0.887 &$24.4\%$& 0.565 & 23.65\\
		\hline
	\end{tabular}\\	
	\label{tab_Results}
\end{table}

Given the proposed suite of trust quantification metrics, we leverage it experimentally to quantify the overall trustworthiness of several well-known, complex deep neural network architectures for the common task of image recognition.  In this case, the set of questions $x$ is a set of images that the oracle asks the deep neural network to recognize, the answers $y$ are the predicted class labels from the deep neural network, the oracle answers $z$ are the true class labels, and the associated confidences $C(y|x)$ for the answers $y$ are the softmax outputs associated with the predicted class labels.  Note that other confidence representations have been proposed in research literature~\cite{Naeini,Zadrozny2002,NiculescuMizil,guo2017calibration}, and can be leveraged within the proposed suite of trust quantification metrics given the flexibility of the proposed trust quantification framework. Therefore, it would be very interesting to investigate their efficacy within the proposed trust quantification framework in the future.  A subset of the ImageNet~\cite{ImageNet} benchmark dataset was used as a set of test questions used in this study to evaluate the trustworthiness of the tested deep neural networks.  More specifically, we leveraged a subset of natural images consisting of 2500 different natural images from the ImageNet dataset.  The tested deep neural networks that are evaluated for trust quantification in this study are four well-known deep image recognition networks that are the most commonly used in the field: 1) MobileNet-V1~\cite{MobileNetv1}, 2) MobileNet-V2~\cite{MobileNetv2}, 3) ShuffleNet-V2~\cite{ShuffleNetv2}, and 4) ResNet-50~\cite{ResNet50}.  For each tested network, we compute the following trust quantification metrics for comparing overall trustworthiness of deep neural networks: 1) NetTrustScore ($T_M$), 2) expected confidence for correct answers (${\bar C}_{y = z}$), and 3) expected confidence for incorrect answers (${\bar C}_{y \neq z}$).  Furthermore, the percentage of correct answers ($n_{y=z}$) and percentage of incorrect answers ($n_{y \neq z}$), along with the architectural complexity of the network in the form of number of parameters are also provided for context.  In addition, we compute the trust spectra for the tested deep neural networks as well as trust densities for specific answer scenarios for a more detailed assessment of where trust breaks down at different levels of granularity.

\subsection{Comparing Overall Trustworthiness}
Table~\ref{tab_Results} shows the NetTrustScore ($T_M$), expected confidence for correct answers (${\bar C}_{y = z}$), expected confidence for incorrect answers (${\bar C}_{y \neq z}$), the percentage of correct answers ($n_{y=z}$), percentage of incorrect answers ($n_{y \neq z}$), and number of parameters for the five tested deep image recognition networks.  A number of interesting observations can be made about the tested deep neural networks based on the trust quantification metrics for comparing overall trustworthiness amongst the networks.

First, it can be observed that of the tested deep neural networks, ResNet-50 achieved the highest NetTrustScore.  This can be attributed to the fact that not only does it get a significantly higher number of answers correct, but also is very confident about these correct answers.  However, it can also be observed that despite ResNet-50 answering significantly more answers correctly than MobileNet-V2 ($\sim$7\%), its NetTrustScore is only $\sim$3.7\% higher than MobileNet-V2.  The reason that the trust gap between ResNet-50 and MobileNet-V2 is not larger despite the large gap in correctness is due to the fact that ResNet-50 is significantly more overconfident than MobileNet-V2 when giving incorrect answers ($\sim$7.2\%), and as such reduces the trustworthiness gap between the two deep neural networks.

Next, it can be observed that for all tested deep neural networks, the expected confidence for incorrect answers is significantly lower than that for correct answers.  This observation is expected and highly desired, as deep neural networks should be hesitant when it is not sure about its answers (which should be the case for incorrect answers) while confident when it is sure (which should be the case for correct answers).  More specifically, it can also be observed that of the tested methods, MobileNet-V2 exhibits the greatest disparity between its expected confidence for correct answers and that for incorrect answers, with its expected confidence for incorrect answers being the only one amongst the different deep neural networks to dip below the 50\% mark.  These are desirable properties from a trust perspective, given that MobileNet-V2 expresses the lowest confidence when it comes to answers it is unsure about and expresses high confidence when it comes to answers it is quite sure are correct.  Therefore, MobileNet-V2 avoids overconfidence and overcautious issues that hinder trust, and achieves a good NetTrustScore despite having very low architectural complexity.

Finally, it can be observed that despite the group of tested deep neural networks that are designed around efficiency and low architectural complexity (i.e., MobileNet-V1, MobileNet-V2, and ShuffleNet-V2) giving significantly lower number of correct answers, the trustworthiness of such networks is quite comparable to networks that are designed around accuracy such as ResNet-50.  For example, despite a huge $\sim$11\% difference in correct answers between ResNet-50 and MobileNet-V1, the difference in NetTrustScore is just $\sim$6.3\%.  Furthermore, despite being significant less architecturally complex in terms of number of parameters, ShuffleNet-V2 achieves a noticeably higher NetTrustScore than MobileNet-V1 (which has $\sim$2.5$\times$ more parameters than ShuffleNet-V2), and has a noticeably larger disparity between its expected confidence for correct answers and that for incorrect answers.  Therefore, based on this observation, there are merits for exploring the design of deep neural networks that can achieve a better balance between complexity, accuracy, and trustworthiness.

\subsection{Analyzing Where Trust Breaks Down}

\begin{figure}
\centering
	\includegraphics[width = 1\linewidth]{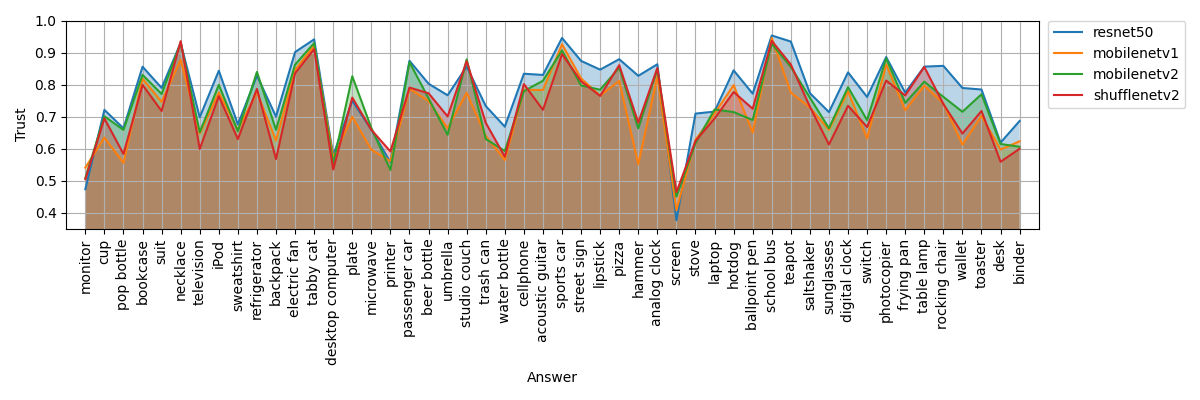}
	\caption{The trust spectra for the four tested deep neural networks.  The trust spectra differences highlight where trust breaks down for specific answers (e.g., answers ``screen'' and ``monitor'' have very low trust while answer ``teapot'' has very high trust variance amongst the tested networks, which we will dive deeper into by studying answer-level trust densities for the different tested networks) can provide great insights into how the networks can be improved to increase trustworthiness.}	
	\label{fig:spectrum}
\vspace{-0.21 cm}
\end{figure}

To get a better understanding of where trust breaks down for the different tested deep neural networks in a detailed manner at different levels of granularity, we will now compare their trust spectra as well as study the trust densities for several interesting answer scenarios.

\subsubsection{Analyzing Trust Spectra}
Figure~\ref{fig:spectrum} shows the trust spectra for the four tested deep neural networks.  A number of interesting observations can be made.  First, it can be observed that there are a number of answers where the trust across all tested deep neural networks is very low (e.g., answer scenarios ``screen'' and ``monitor'', two scenarios that all deep neural networks are confused about and have difficulty answering with great confidence).  Seeing such outliers of low trust indicates that all four of the tested deep neural networks provide unreliable answers under such answer scenarios, which would warrant deep investigations in real-world production scenarios.  Second, it can be observed that there are a number of answer scenarios where the trust variance amongst the different deep neural networks is very high (e.g., answer ``teapot''), which indicates that certain deep neural networks tested in this study are potentially much more reliable than others in particular answer scenarios.  As such, the trust spectra differences highlight where trust breaks down at the network level and can provide great insights into how the networks can be improved to increase trustworthiness.

\begin{figure}[t]
    \centering
    \begin{tabular}{cc}
        \subfloat[Pop Bottle\label{fig:pop_bottle_density}]{\includegraphics[width=0.47\linewidth]{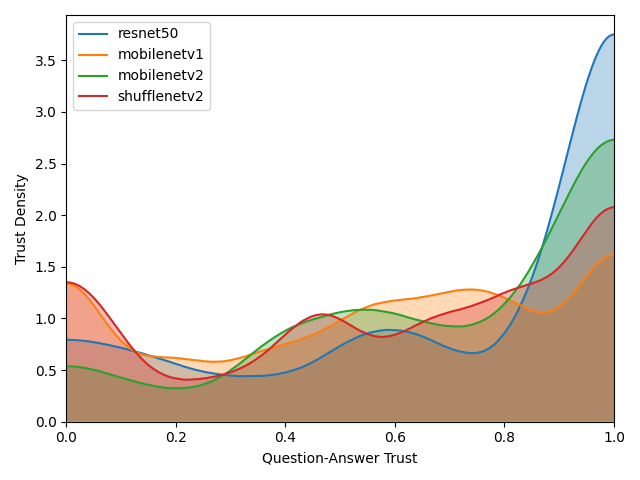}} &
        \subfloat[Hammer\label{fig:hammer_density}]{\includegraphics[width=0.47\linewidth]{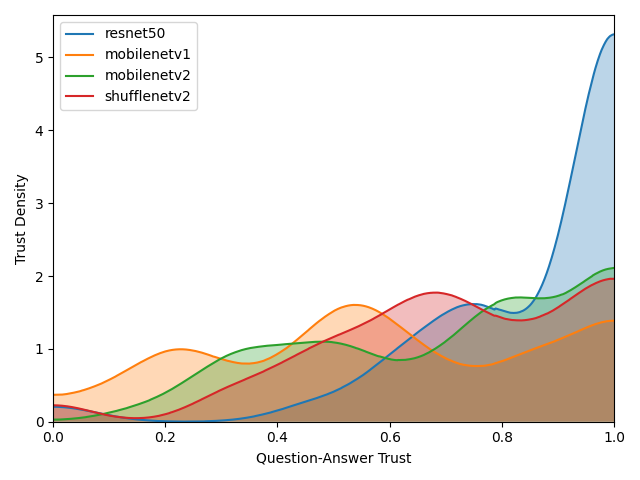}} \\
        \subfloat[Screen\label{fig:screen_density}]{\includegraphics[width=0.47\linewidth]{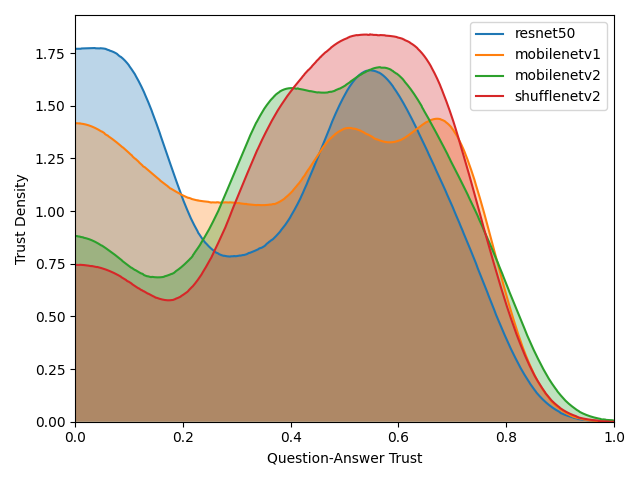}} &
        \subfloat[Teapot\label{fig:teapot_density}]{\includegraphics[width=0.47\linewidth]{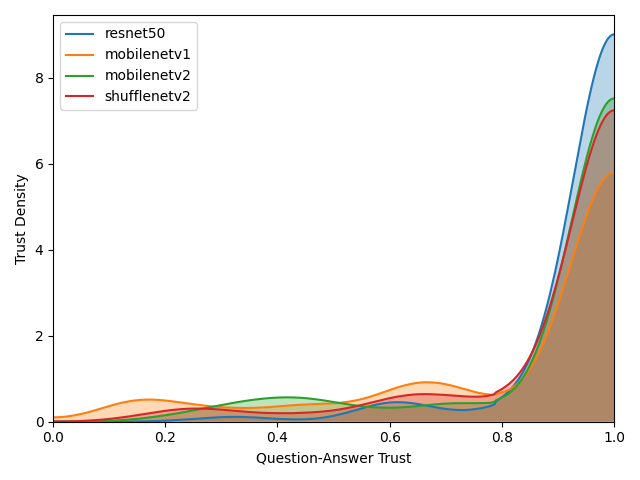}}

    \end{tabular}
    \caption{The trust densities for four different answer scenarios in a subset of the ImageNet dataset: (a) ``Pop bottle'', (b) ``Hammer'', (c) ``Screen'', and (d) ``Teapot''.}
    \label{fig:trustdensity_cases}
\end{figure}

\subsubsection{Analyzing Trust Densities}
We now take a deeper dive into where trust breaks down at the answer level by studying the trust densities of the tested deep neural network at a number of interesting answer scenarios. Figure~\ref{fig:pop_bottle_density} shows the trust density for answer scenario ``pop bottle''. In this scenario, all four tested deep neural networks demonstrate large trust variances across all questions that are answered as ``pop bottle''. Not only is there significant variance in the overall trust for the ``pop bottle'' answer scenario across the tested deep neural networks as seen in the trust spectra (ranging from 0.56 to 0.66), the even higher trust variance exhibited in the trust density makes it difficult to judge which of the deep neural networks is more trustworthy compared here.  The defining difference between the tested deep neural networks in this answer scenario can be seen in the densities in the high trust regions (the top right of Figure~\ref{fig:pop_bottle_density}). It is in these high trust regions that ResNet-50 has a distinct advantage at producing more trustworthy answers for the ``pop bottle'' answer scenario.

Figure~\ref{fig:hammer_density} shows trust densities for the ``hammer'' answer scenario, where the trust spectrum coefficient differences between the tested deep neural networks based on their trust spectra is at its largest. Here, it is very clear that ResNet-50 produces significantly more trustworthy answers for the ``hammer'' answer scenario compared to the other tested deep neural networks, with a significant skew towards high trust regions.  Interestingly, two of the deep neural networks (MobileNet-V2 and ShuffleNet-V2) have almost identical trust spectrum coefficients for this particular answer scenario but have drastically different trust densities.  Both models have multi-modal behaviours, where MobileNet-V2's trust density here exhibit higher densities at both extremes, and ShuffleNet-V2's trust density exhibit higher density in the mid-range trust region and a moderate amount of density in the higher trust range. Such detailed behavioural knowledge about the trustworthiness of deep neural networks at the answer level can be useful for better understanding where trust breaks down and for analyzing whether methods for mitigating trust issues are in fact having appropriate impact on behaviour.

Figure~\ref{fig:screen_density} shows the trust densities for the ``screen'' answer scenario where all tested deep neural networks exhibit low overall trust as seen in the trust spectra. In such a scenario, it is important to identify why the deep neural networks are producing such untrustworthy answers. For example, the trust density of ResNet-50 for this answer scenario exhibits a distinct bimodal distribution. The bimodal distribution of ResNet-50's trust density is particularly interesting in this case because of how untrustworthy some of the predictions are for this particular scenario, despite achieving the highest NetTrustScore when compared to the other tested deep neural networks.  This observation illustrates the importance of analyzing trust at different scales (in this case, at the coarse-grain level with NetTrustScore, at the intermediate level with trust spectrum, and at the lowest level with trust density) to get a more complete trust profile of a deep neural network.  This observation also provides a clear direction for improving the trustworthiness of ResNet-50 under this particular answer scenario. Analyzing commonalities between the strongly untrustworthy question-answer pairs near the low trust peak of the ResNet-50's trust density in this particular answer scenario would likely provide important insight into why ResNet-50's answers are so unreliable here, and would allow a deep learning practitioner to remedy the deep neural network's (or dataset's) impaired performance.

Finally, Figure~\ref{fig:teapot_density} shows the trust densities for the ``teapot'' answer scenario, another scenario where there is high overall trust variance across tested deep neural networks according to the trust spectra, but this time the overall trust for all networks is on the higher end of the spectrum.  The high overall trust for all networks for this particular scenario indicates that the deep neural networks are independently trustworthy here. An interesting trend observed from the trust densities is that the lower the trust density variance is for a given deep neural network, the higher the peak is for that network in the very high trust regions, and by extension the higher the overall trust for that particular answer scenario. Such a trend reinforces the credibility and trustworthiness of the each tested deep neural networks to be trustworthy for the ``teapot'' answer scenario.

As can be seen from the deep dive analysis into these four different answer scenarios, identifying answer scenarios of concern or of interest by analyzing the trust spectrum and then studying the answer-level trust densities of these answer scenarios of interest can enable a much deep understanding and gain valuable insights into where trust breaks down and potential directions to find the cause of such trust breakdowns.

\section{Conclusions}
\label{Conclusions}

In this study, we introduced a suite of quantitative metrics for assessing the overall trustworthiness of deep neural networks based on their behaviour when answering a set of questions.  The proposed question-answer trust, trust density, trust spectrum, and NetTrustScore are simple to calculate and easy to interpret, and aligns with past social psychology studies studying the relationship between trust and confidence in such a scenario.  The proposed NetTrustScore was leveraged to quantify the overall trustworthiness of several well-known, complex deep neural network architectures for the common task of image recognition using a subset of the ImageNet dataset, and led to interesting insights into the behaviour of deep neural networks from a trust perspective when giving both correct and incorrect answers. A comparative analysis of the trust spectra across deep neural networks was further conducted and an analysis of trust densities for particular answer scenarios was performed to get a better understanding of where trust may break down at different levels of granularity.

Future work involves exploring the extension of the proposed metrics to quantify the overall trustworthiness of deep neural networks for unsupervised learning scenarios, as well as experimentally exploring the overall trustworthiness of deep neural networks designed for other types of complex tasks such as natural language processing, object detection, speech recognition, trend analysis, predictive recommendation, and times-series predictive analysis, given that such tasks are highly relevant to real-world applications where trustworthiness is critical for widespread deployment in society.  Furthermore, a number of other confidence representations outside of the one leveraged in this study have been proposed in research literature~\cite{Naeini,Zadrozny2002,NiculescuMizil,guo2017calibration}, many of which leverage different calibration strategies to produce confidence values that better reflect the true correctness likelihood.  Given that these confidence representations can be leveraged within the proposed suite of trust quantification metrics due to the flexibility of the proposed trust quantification framework, it would be very interesting to investigate their efficacy and their behaviour within the proposed trust quantification framework across a variety of different deep neural network architectures.  Finally, we aim to explore in detail ways to better quantify and analyze where trust break downs in the various decisions that deep neural networks make beyond what trust spectrum and trust density provides, which will give much better understanding on how to further improve and address trust in deep neural networks.

\section{Broader Impact}
\label{impact}
In the earlier days of artificial intelligence, the primary focus of practitioners in the field has been largely aimed at the notion of performance and function.  This focus around performance and function made logical sense because at that early stage, just getting artificial intelligence algorithms to behave and make reasonable predictions and decisions for complex, abstract tasks was considered a tremendous struggle that was often intractable or impractical given the computing power and algorithmic constructs at the time.  However, with the recent tremendous advances  machine learning, particularly deep learning, demonstrating previously unthinkable levels of performance on highly complex tasks such as visual perception, speech recognition, natural language understanding, and even game playing, and the availability of big data and big compute, artificial intelligence is no longer seen as a distant future but a ubiquitous technology now.  Machine learning is on every new smartphone sold, powers the conversational voice assistants that sit directly in our homes, guides the way we navigate the world through intelligent map agents, influences the way we shop, controls what media we consume, holds sway over which candidate receives an interview, and much, much more.

This enormous level of influence that machine learning has over the socioeconomic fabric has now led practitioners as well as regulators to look beyond just performance and functionality in judging the machine learning algorithms that influences our daily lives. Be it accountability, fairness, transparency, ethics, robustness, or many other non-functional related factors, the scrutiny of machine learning when integrated into society has increased greatly as its proliferation increases.  Amongst the key factors upon which machine learning algorithms are now being considered is the notion of trust, and it is under this particular lens that we were motivated when devising the proposed set of metrics for trust quantification of deep neural networks.  The proposed metrics are by no means perfect, but the hope is to push the conversation towards better quantitative metrics for evaluating the overall trustworthiness of deep neural networks to help guide practitioners and regulators in producing, deploying, and certifying deep learning solutions that can be trusted to operate in real-world, mission-critical scenarios.

\bibliographystyle{IEEEtran}
\bibliography{trustscore}

\end{document}